# A two-stage algorithm in evolutionary product unit neural networks for classification

Antonio J. Tallón-Ballesteros [a,*], César Hervás-Martínez [b]

[a] Department of Languages and Computer Systems, University of Seville, Reina Mercedes Avenue, Seville 41012, Spain
[b] Department of Computer Science and Numerical Analysis, University of Córdoba, Campus of Rabanales, Albert Einstein Building, Córdoba 14071, Spain

ABSTRACT

This paper presents a procedure to add broader diversity at the beginning of the evolutionary process. It consists of creating two initial populations with different parameter settings, evolving them for a small number of generations, selecting the best individuals from each population in the same proportion and combining them to constitute a new initial population. At this point the main loop of an evolutionary algorithm is applied to the new population. The results show that our proposal considerably improves both the efficiency of previous methodologies and also, significantly, their efficacy in most of the data sets. We have carried out our experimentation on twelve data sets from the UCI repository and two complex real-world problems which differ in their number of instances, features and classes.

## 1. Introduction

Maintaining diversity in early generations is a crucial task in evolutionary algorithms (EAs) because it can affect not only the convergence speed but also the quality of the final solution. A diverse population is preferable at the beginning of the algorithm and a more condensed one at the end of the search (Maaranen, Miettinen, & Mäkelä, 2004). However, to date the issue of population initialization has received surprisingly little attention in EA literature while, on the other hand, the topic of operators and the representation of individuals has been amply covered. On a lower abstraction level, we must think about how the individuals that constitute the population are generated. In most cases, as in this paper, in evolutionary computation (EC) a pseudo-random number generator is employed, although it is often expressed as "the population is generated randomly"; the idea behind this kind of generators is to obtain a set of values that imitates a random sequence (Maaranen, Miettinen, & Penttinen, 2007).

Another question, associated with diversity throughout the evolutionary process, is: does it create only one population or various populations? The issue of generating various populations has also been discussed previously. Wang et al. (Wang, Zheng, & Tang, 2002) present the idea that some populations undergo the application of different mutation operators; next, all the individuals are mixed and split into several populations, each then undergoing the application of yet another mutation operator. In De Garis (1990), an iterative scheme with several fitness functions is proposed, so that the population evolves by means of a GA using the first fitness function, then the resulting population is employed as the current population in a GA that uses the second fitness function and so on. Generally, the populations explore several areas in the search space by means of different seeds. The number of populations may vary and there is no single common accepted value.

In the context of artificial neural networks (ANNs), our proposal, which will be described in detail in continuation, diversifies the architecture of the neural network at the beginning of the evolutionary process. The first stage consists of creating two populations with different features (the maximum number of nodes in the hidden layer, so the topology will be different), evolving them with identical parameter values of the EA, for a small number of generations, selecting the best individuals from each population in the same proportion and combining them to constitute a new population. In the second stage, the main loop of the standard EA used will be applied to the new population. In this way, the population has wider diversity due to the different topologies found in the neural networks. The initial short training leads to random individuals to explore possible promising areas in two directions, since there are two different populations. After that, individuals with different topologies coexist and the more suitable ones will remain. To be coherent with our previous work (Tallón-Ballesteros, Gutiérrez-Peña, & Hervás-Martínez, 2007) we have used two populations in the first phase that will be merged into one. In this way we can better analyse previous and current performances. We will compare the accuracy, complexity and efficiency of two previous experiments with a current one for all the problems under consideration.

* Corresponding author. Tel.: +34 954556237; fax: +34 954557139.
*E-mail addresses:* atallon@us.es (A.J. Tallón-Ballesteros), chervas@uco.es (C. Hervás-Martínez).

The issue of diversity is very important in EC. Evolution, by definition, requires diversity, which refers to the (genetic) variation in population members (Amor & Rettinger, 2005; Heitkoetter & Beasley, 2001). Diversity is valuable since new areas of the search space are explored, without which the search could remain trapped in a local optimum (Chop & Calvert, 2005). Actually a high diversity implies that the population covers a larger part of this space (Ursem, 2002). A common view of the evolutionary process is that diversity enhances the performance of a population by providing more opportunities for evolution. A homogeneous population offers no advantage for improvement as the entire population is focused in a particular portion of the search space. By contrast, a diverse population will simultaneously sample a large area of search space, providing the opportunity to locate different, potentially better, solutions (Curran & O'Riordan, 2006). Diversity can be considered at different stages, basically in the first steps of evolution or throughout the evolutionary phase. This paper focuses on the former, although there are several papers that consider the latter. A great deal of research has established diversity as a target to promote, maintain or reintroduce into evolving populations of solutions in order to achieve maximum performance (Curran & O'Riordan, 2006). In Amor and Rettinger (2005), there are some references to studies referring to the above aspects. As previously mentioned, diversity can be considered in the evolutionary stage. For instance, with respect to the selection stage, rank scaling (Goldberg, 1989) is a sample. In the replacement phase, several proposals have been put forward, like the hybrid replacement scheme proposed by Lozano et al. (Lozano, Herrera, & Cano, 2008) or crowding methods (Mahfoud, 1995). High diversity does not imply better GA performance; this is closely related to the question of exploration versus exploitation, but enforcing diversity in the early phases of evolution ensures a broad exploration of the search space (Amor & Rettinger, 2005).

Our objective is to improve the efficiency (measured by means of the number of evaluations) and efficacy, if possible, of the previous models that have been employed to date by us. The training of databases, which have different numbers of patterns, features and classes, is dealt with by means of ANNs to evaluate the methodology proposed. The computational cost is very high if EAs with different parameter settings are employed for the training of the above-mentioned networks. It is convenient to change the usual way of generating network models and, thus obtain enough diverse ones with respect to the architecture in the first generations of the evolution. So, instead of using the methodology in Martínez-Estudillo, Martínez-Estudillo, Hervás-Martínez, and García-Pedrajas (2006) that consists of generating $10{*}N$ neural networks randomly with a maximum fixed number of nodes in the hidden layer, sorting them according to their fitness and choosing the best $N$ ones to constitute the initial population, we present the idea of creating two populations. Each population has a different maximum number of nodes in the hidden layer, and is evolved for a small number of generations in order to subsequently merge the best half of individuals of each one into a single population; then we apply the standard evolution process of the EA employed to date for classification by means of product unit neural networks (PUNN) (Martínez-Estudillo, Hervás-Martínez, Gutiérrez-Peña, Martínez-Estudillo, & Ventura-Soto, 2006).

In this way, the training time will be reduced because the population will be much more diverse. Moreover, it will not be necessary to apply the full evolutionary process to some populations with different numbers of nodes in the hidden layer. Only there is more than one population in the first step of the evolution; next we merge individuals with different topologies into one. This methodology is more recommendable in the case of data sets with a great number of patterns, given that the processing time is very high for a complete configuration with a reasonable number of iterations for a proper number of generations. Logically, it is not usual to carry out experiments with a single configuration but with a number of them, resulting in a computationally-intensive procedure. The advantage of our proposal is that all individuals do not present a fixed architecture, but a flexible topology between two possible ones. The performance will not be so dependent on parameter tuning, in the sense that the maximum number of hidden nodes in an individual can fluctuate between two user-defined values over the course of the evolution.

Several runs of the algorithm have been performed to smooth the stochastic character of the EAs using mean values in order to complete a statistical analysis of the results obtained. After presenting the results and the number of EA evaluations with both earlier and current methodologies, there will be an analysis to determine whether the performance of the solutions improves quality-wise with respect to the Correct Classification Ratio (CCR) obtained with our previous methodology, and if the complexity with respect to the number of connections does as well. In order to do this, statistical tests will be used to compare the mean performances of the CCR and the mean number of connections obtained with both the proposed methodology and the standard method. Finally, there will be a comparison of both methodologies, that use PUNN, and other techniques that employ neural networks based on sigmoidal and radial basis functions. There is also a general review of the results obtained with other neural networks or classical/modern machine learning approaches.

This paper is organized as follows: Section 2 describes some concepts about PUNN and the EA; Section 3 presents the description of our proposal; Section 4 details the experimentation process; then Section 5 shows and analyzes the results obtained; finally, Section 6 states the concluding remarks.

## 2. Methodology

### 2.1. Product unit neural networks

Many different types of neural network architectures have been used, but the most popular one has been the single-hidden layer feed-forward network. Amongst the numerous algorithms for training neural networks in classification problems, our attention focuses on evolutionary artificial neural networks (EANNs). EANNs have been a key research area in the past decade providing an improved platform for optimizing network performance and architecture (number of hidden nodes and number of connections) simultaneously.

Up to now, designing topologies is still very much a human expert's job. It depends heavily on expert experience and a tedious trial-and-error process. There is no systematic way to automatically design a near-optimal architecture for a given task (Yao, 1999). Design of the optimal architecture for an ANN can be formulated as a search problem in the architecture space where each point represents one. Given some performance (optimality) criteria, e.g., lowest training error, lowest network complexity based on the number of connections, etc., about topologies, the performance level of all of them forms a discrete surface in the space. The optimal topology design is equivalent to finding the highest point on this surface (Yao & Liu, 1997).

Miller, Toddm, and Hegde (1989) proposed that EC was a very good candidate for searching the space of architectures because the fitness function associated with that space is complex, noisy, non-differentiable, multi-modal and deceptive. Since then, many evolutionary programming methods have been developed to evolve ANNs, for instance, those by Yao and Liu (1997) and Yao (1999).

The methodology employed here consists of the use of an EA as a tool for learning the architecture and weights of a PUNN model

(Martínez-Estudillo, Hervás-Martínez, et al., 2006; Martínez-Estudillo, Martínez-Estudillo, et al., 2006). This class of multiplicative neural networks comprises such types as sigma-pi networks and product unit networks. Some of the advantages of PUNN are increased information capacity and the ability to form higher-order combinations of inputs (Durbin & Rumelhart, 1989). Besides that, it is possible to obtain upper bounds of the Vapnik–Chervonenkis dimension of PUNN similar to those obtained for sigmoidal neural networks (Schmitt, 2001). Finally, it is a straightforward consequence of the Stone-Weierstrass Theorem to prove that PUNN are universal approximators (Martínez-Estudillo, Martínez-Estudillo, et al., 2006). Despite these advantages, product unit based networks have a major drawback. Networks based on product units (PUs) have both more local minima and more probability of becoming trapped in them (Ismail & Engelbrecht, 2000). The main reason for this difficulty is that small variations in the exponents can cause large changes in the total error surface.

Several efforts have been made to carry out learning methods for PUs. Janson and Frenzel (1993) developed a GA for evolving the weights of a network based on PUs with a predefined architecture. The major problem with this kind of algorithm is how to obtain the optimal architecture before-hand (Ismail & Engelbrecht, 2000). Unfortunately, up to the present, the problem of designing a near optimal ANN architecture for an application remains unsolved. Abraham (2004) presents MLEANN (meta-learning EANN), an adaptative computational framework based on evolutionary computation for automatic design of optimal ANNs. However, this paper defines two types of experiments, and it is necessary to supply the number of hidden nodes as a user-specified parameter. The first one is more flexible, in the sense that the topology is indicated with a number of hidden nodes that ranges between 5 and 16. The disadvantage of using an EA in the training of PUNN is that the processing time could be too great with respect to the dimension of the features' space and to the number of classes in a concrete classification problem under consideration. Thus, our proposal tries to ease and speed up the way of finding a good topology regarding previous works. So, it does not define a fixed architecture, but one that combines individuals that present two different values of hidden nodes.

Fig. 1 shows the structure of a PUNN model for a bi-classification problem; this is a three-layer architecture, that is, $k$ nodes in the input layer, m ones and a bias one in the hidden layer and one node in the output layer. The topologies will be indicated by means of the numbers of nodes in each layer from input to output given as a sequence: *Number of inputs: number of nodes in the hidden layer: number of nodes in the output layer*. So, in Fig. 1 we have a *k:m:1* architecture.

The transfer function of each node in the hidden and output layers is the identity function. Thus, the functional model obtained by each of the nodes in the output layer is given by:

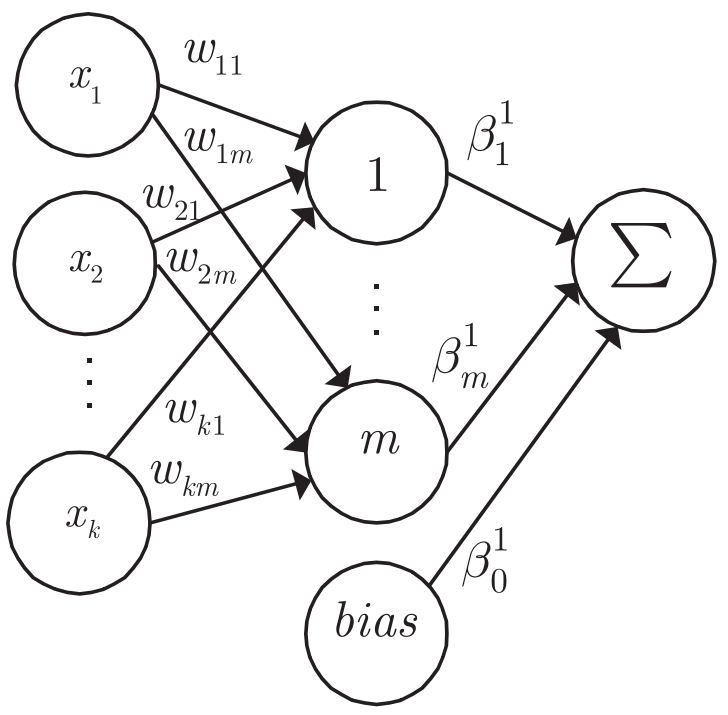

**Fig. 1.** Structure of a PUNN model for a bi-classification problem.

$$f(x_1, x_2, \ldots, x_k) = \sum_{j=1}^{m} \beta_j^l \left( \prod_{i=1}^{k} x_i^{w_{ji}} \right) + \beta_0^l. \quad (1)$$

## 2.2. Evolutionary algorithm

We use an EA to design the structure and learn the weights of PUNN. The search begins with a random initial population and, for each iteration, the population is modified using a population-update algorithm. The population is subjected to the operations of replication and mutation. Crossover is not used due to its potential disadvantages in evolving artificial networks (Angeline, Saunders, & Pollack, 1994; Yao & Liu, 1997). With these features the algorithm falls into the class of evolutionary programming. We have a classification problem and the general scheme of the EA, depicted in Fig. 2, is the following:

Next, we are going to explain the main aspects of the EA:

(1) Notation, data, input and output parameters and variables
We have indicated the keywords in boldface and the functions in italics. Comments are preceded by a double-slash (//). The necessary data is a training set with the instances of a classification problem. The main parameters of the EA are the maximum number of generations (*gen*) and the number of nodes in the hidden layer (*neu*). The remaining parameters will be described further on. At the end of the EA, it returns the best PUNN model with *neu* nodes in the hidden layer. The variables used are the number of the current generation ($t$), the last generation (*last_generation*) and two arrays, one with the individuals of the evolving population and the other with the fitness of each individual.

(2) Representation of the individuals
Regarding the representation of individuals, the EA treats the population like a set of PUNN models. An object-oriented approach has been adopted and the algorithm deals directly with the ANN phenotype. Each connection is specified by a binary value indicating if the connection exists, and a real value representing its weight. As the crossover is not considered, this object-oriented representation does not assume a fixed order between any hidden nodes. All the individuals in the population have the same maximum number of neurons in the hidden layer. The concrete value must be indicated as an EA input and this will clearly affect the performance and complexity of the neural network. Obtaining the optimal value is a challenge.

(3) Error and fitness functions
We have considered a standard softmax activation function, associated with the $g$ network model, given by:

$$g_j(\mathbf{x}) = \frac{\exp f_j(\mathbf{x})}{\sum_{i=1}^{L} \exp f_i(\mathbf{x})} \quad j = 1, \ldots, L, \quad (2)$$

where $L$ is the number of classes in the problem, $f_j(\mathbf{x})$ is the output of node j for pattern $\mathbf{x}$ and $g_j(\mathbf{x})$ is the probability that this pattern belongs to class $j$. Taking this into account, a function of cross-entropy error is used to evaluate a network $g$ with the instances of a problem, which is reflected in the following expression:

$$l(g) = -\frac{1}{N} \cdot \sum_{i=1}^{N} \sum_{j=1}^{L} \left( y_i^j \ln(g_j(\mathbf{x}_i)) \right) \quad (3)$$

and substituting $g_j$ defined in (2),

$$l(g) = \frac{1}{N} \cdot \sum_{i=1}^{N} \left( -\sum_{j=1}^{L} y_i^j f_j(\mathbf{x}_i) + \ln \left( \sum_{j=1}^{L} \exp f_j(\mathbf{x}_i) \right) \right), \quad (4)$$

```
Program: Evolutionary Algorithm
Data: Training set
Input parameters: gen, neu
Output: Best PUNN model
1:  t ← 0
2:  P(t) ← {ind_1, …, ind_10000}                              // Random initialisation of the population
3:  f(P(t) {ind_1, …, ind_10000}) ← fitness (P(t) {ind_1, …, ind_10000})   // Calculate fitness
4:  P(t) ← P(t) {ind_1, …, ind_10000}                         // Sort individuals by fitness: ind_i > ind_i+1
5:  P(t) ← P(t) {ind_1,… ind_1000}                            // Retain the 1000 best ones
6:  while stop criterion not met do                           // main loop
7:      P(t) {ind_901,…ind_1000} ← P(t) {ind_1, …, ind_100}   // Best 10% replace the worst 10%
8:      P(t+1) ← P(t) {ind_1, …, ind_900}
9:      P(t+1) ← pm (P(t+1) {ind_1, …, ind_90})               // Parametric mutation (10% P(t+1))
10:     P(t+1) ← sm (P(t+1) {ind_91, …, ind_900})             // Structural mutation (90% P(t+1))
11:     f(P(t+1) {ind_1, .. ind_900}) ← fitness (P(t+1) {ind_1, …, ind_900})   // Evaluate
12:     P(t+1) ← P(t+1) (ind_1, …, ind_900) ∪ P(t) {ind_901, …, ind_1000}
13:     P(t+1) ← P(t+1){ind_1, …, ind_1000}                   // Sort individuals
14:     t ← t+1
15:     last_generation ← t+1
16: end while
17: return best (P(last_generation) {ind_1})
```

**Fig. 2.** Pseudocode of the EA.

where $y_i^j$ is the target value for class $j$ with pattern $\mathbf{x}_i$ ($y_i^j = 1$ if $\mathbf{x}_i \in$ class $j$ and $y_i^j = 0$ otherwise), $f_j(\mathbf{x}_i)$ is the output value of the neural network for the output neuron $j$ with pattern $\mathbf{x}_i$, $N$ the number of patterns and $L$ the number of classes. Observe that softmax transformation produces positive estimates that sum to one and therefore the outputs can be interpreted as the conditional probability of class membership. On the other hand, the probability for one of the classes does not need to be estimated because of the normalization condition. Usually, one activation function is set to zero; in this work $f_L(\mathbf{x}_i) = 0$ and we reduce the number of parameters to estimate. Thus, the number of nodes in the output layer is equal to the number of classes minus one in the problem.

Since the EA objective is to minimise the chosen error function, a fitness function is used in the form $A(g) = (1 + l(g))^{-1}$.

(4) Initialisation of the population

At the beginning of the EA, $10*N$ individuals are generated randomly (step 2) by means of a pseudo-random number generator, $N$ being the population size; in the current paper it is equal to 1000. Next, all individuals are evaluated, sorted by decreasing fitness and the best $N$ ones will compose the initial population (steps 3–5).

(5) Stop condition

The main loop of the EA is repeated until the maximum number of generations (*gen*) is reached or until the best individual or the population mean fitness do not improve during *gen-without-improving* generations (20 in this paper).

(6) Parametric mutation

Parametric mutation changes the value of the model coefficients (step 9) and consists of a simulated annealing algorithm. The severity of a mutation of an individual $g$ in the population is dictated by the temperature $T(g)$, given by $T(g) = 1 - A(g)$, $0 \leqslant T(g) < 1$. Parametric mutation is accomplished for each exponent $w_{ji}$ and coefficient $\beta_j^l$ of the model with Gaussian noise, where the variance depends on the temperature:

$$w_{ji}(t+1) = w_{ji}(t) + \xi_1(t) \quad j = 1, \ldots, k \quad i = 1, \ldots, m, \tag{5}$$

$$\beta_j^l(t+1) = \beta_j^l(t) + \xi_2(t) \quad j = 0, \ldots, m \quad l = 0, \ldots, L-1, \tag{6}$$

where $\xi_k(t) \in N(0, \alpha_k T(g))$ $k = 1, 2$, represents a one-dimensional normally distributed random variable with mean 0 and variance $\alpha_k(t) \cdot T(g)$, and $t$ is the $t$-th generation. It should be pointed out that the modification of the exponents $w_{ji}$ is different from the change of the coefficients $\beta_j^l$, therefore $\alpha_1 \ll \alpha_2$. The effect of a mutation on the weight from an input-variable to a hidden node is greater than from a hidden node to an output node, so the changes in exponents $w_{ji}$ should be smaller than in the coefficients $\beta_j^l$. Since $\alpha_2$ acts on the coefficients of the output layer, this parameter controls the diversity of the individuals in the population. $\alpha_2$ is multiplied by the network temperature, so at the beginning there is a high temperature to be able to move from one solution to another with a very different fitness, but in the end the temperature is low, so with $\alpha_2$ higher values, greater diversity can be achieved. With a $\alpha_1$ high-value, the algorithm may be made to reach a premature convergence. In our case, since this is a classification problem, the evolution process must be short. That is why an evolutionary mechanism must be selected for parameters $\alpha_1$ and $\alpha_2$ that converge toward optimum values more quickly. Rechenberg's 1/5 success rule has been applied (Rechenberg, 1973).

(7) Structural mutation

This implies a modification in the structure of the model (step 10) and allows different regions in the search space to be explored while helping to maintain the diversity of the population. There are five different structural mutations, the first four ones are similar to those in the GNARL model (Angeline et al., 1994): node addition, node deletion, connection addition, connection deletion and node fusion. All the above mutations are made sequentially in the given order, with probability $T(g)$, in the same generation on the same network. If probability does not select a mutation, one of the mutations is chosen at random and applied to the network.

(8) Summary of the parameters of the EA

To complete the specification of the EA, we will now explain some EA parameters or features. As stated above, the values of specific parameters like the maximum number of generations *(gen)*, the number of neurons in the hidden layer (*neu*) must be indicated in the EA as input values. There are no typical values for them, so the difficulty lies in determining good values. Also, the performance of the EA depends above all on these values. Finally, to conclude this section Table 1 describes the values of some general EA parameters.

## 3. Proposal description

In Tallón-Ballesteros et al. (2007), we proposed an experimental design distribution (EDD) that will be our starting point here. This consists of distributing some parameters, either of the network topology or of the EA, as the number of nodes in the hidden layer, the number of generations and the output-variance value ($\alpha_2$), over some computing nodes; each set of concrete values of previous parameters is called a configuration. To do this, an initial configuration, called the base configuration, is defined and it is modified with new values in one/two parameters in each of the computing nodes. Thus, once the modifications have been made, each of the processing nodes will run the EA with a different configuration. Proposals were presented to distribute two/three parameters, although this paper continues in the line of distributing three parameters. In our previous study, there were eight different configurations of the EA working with the same data set, four of them undergoing long training while the remaining ones were shorter.

**Table 1**
General EA parameters/features.

| Parameter/Feature | Value |
|---|---|
| Population size ($N$) | 1000 |
| *gen-without-improving* | 20 |
| Interval for the exponents $w_{ji}$/ coefficients $\beta_j^l$ | [−5,5] |
| Initial values of $\alpha_1$ and $\alpha_2$ | 0.5 and 1 respectively |
| Normalization of the input data | [1,2] |
| Number of nodes in node addition and node deletion operators | [1,2] |

**Table 2**
Description of the EDD configurations.

| Configuration | Num. of neurons (*neu*) | Max. Num. of generations | $\alpha_2$ |
|---|---|---|---|
| 1 | neu | *gen* | 1 |
| 2 | neu+1 | *gen* | 1 |
| 3 | neu | *gen* | 1.5 |
| 4 | neu+1 | *gen* | 1.5 |

At present, as our main goal is to improve efficiency without losing efficacy, our attention is focused on the experiments with a long training time. It is equivalent to saying that our interest now lies in distributing two parameters, *neu* and $\alpha_2$. This decision was adopted because these parameters have a great impact on the performance of the classifier. Table 2 presents the description of the EDD configurations related to this paper. In these configurations the *gen* parameter takes the value indicated as the input to the EA. In this situation four different EA configurations are run, combining two different values for each of the parameters.

The current paper presents a procedure called the Two-Stage Evolutionary Algorithm (TSEA). First of all, a number of neurons is fixed in the hidden layer, *neu*. The first stage consists of generating two populations of size $N$, respectively with *neu* and *neu*+1 nodes. These populations evolve for 0.1**gen* generations. Afterwards, the best $N/2$ individuals in each population are selected and merged into a new population of size $N$. In the second stage, the main loop of the standard EA is applied to the new population. The parameters are defined as $N$, the size of the population; *gen*, the maximum number of generations; and *neu*, the maximum number of nodes in the hidden layer. The values of these parameters are explained further on. Figs. 3 and 4 present the TSEA scheme and pseudocode.

In the previous model, two full independent experiments (steps a) and (b) of Fig. 5 had to be run, one with each of the architectures, for *gen* generations. With the current proposal, however, each of the architectures only has to be run for a few generations (0.1**gen*) followed by a full evolutionary cycle to a new population (step c) that combines individuals from both topologies. Concerning the number of generations, previously 2**gen* were needed with EDD, while now only 1.2**gen* is needed with TSEA.

The features of TSEA are the following:

- PUNN have been employed with: a number of neurons in the input layer equal to the number of variables in the problem; a hidden layer with a number of nodes that depends on the data set to be classified; and the number of nodes in the output layer equal to the number of classes minus one because a softmax-type probabilistic approach has been used.
- Two populations have been generated with 1000 individuals in each of the experiments.
- Each of the populations is evolved for 0.1**gen* generations, whose concrete value depends on the data set. Once this short evolutionary process has been carried out, the best 500 individuals in each population are selected and are merged to constitute the new population numbering 1000. This new population will evolve for *gen* generations.
- Two experiments have been performed for each problem, where two different values have been used for $\alpha_2$, associated with the residual of the updating expression of the output layer weights. Thus, two different configurations have been considered. Table 3 describes them and the values employed for each of the most relevant parameters.

## 4. Experimentation

### 4.1. Data sets

Table 4 summarizes the data sets employed. Most of them are publicly available at the UCI repository (Asuncion & Newman,

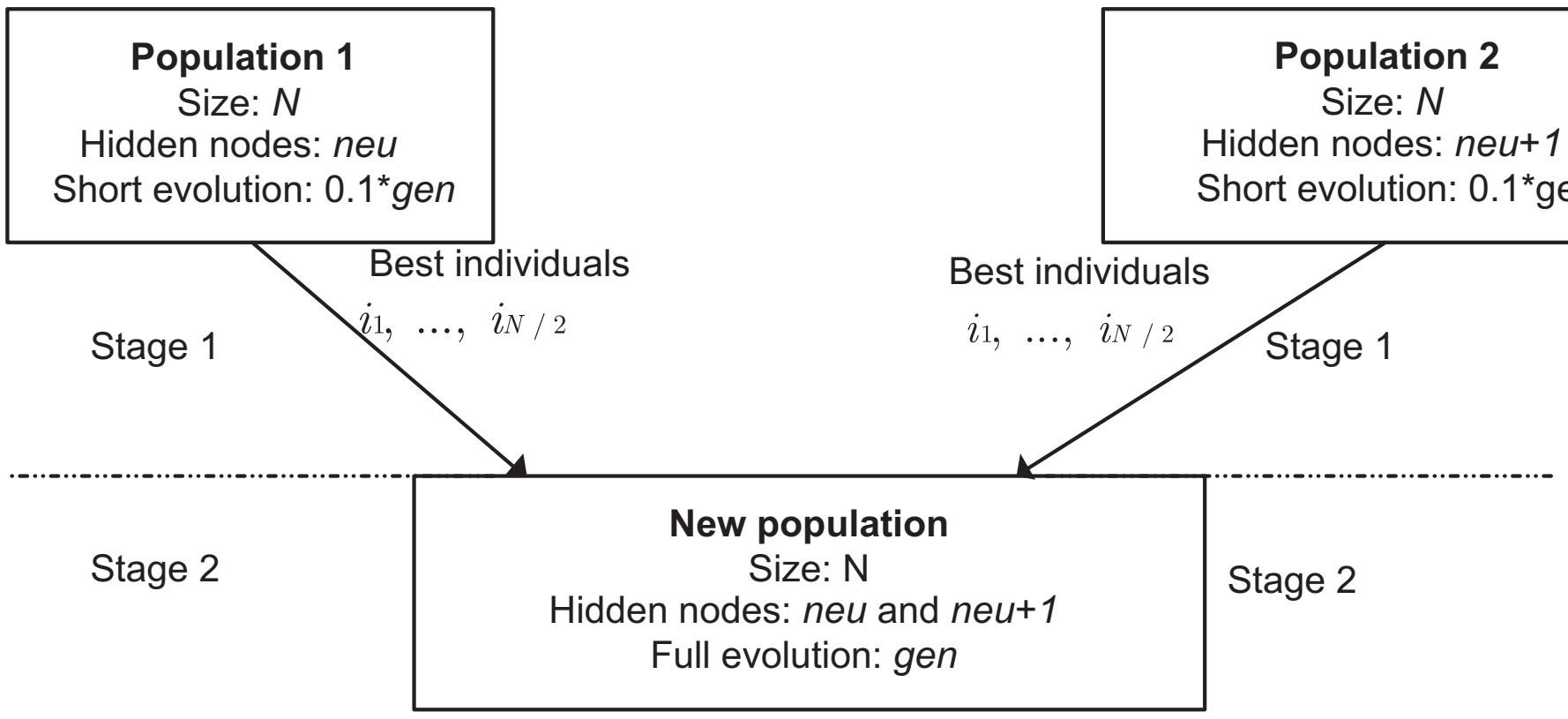

**Fig. 3.** Scheme of the TSEA.

```
Program: Two-Stage Evolutionary Algorithm
Data: Training set
Input parameters: gen, neu
Output: Best PUNN model
                                   // First Stage
t ← 0
                                   // Population P_1
P_1(t) ← {ind_1, …, ind_10000}              // Individuals of P_1 have neu nodes in the hidden layer
f_1 (P_1(t) {ind_1, …, ind_10000}) ← fitness (P_1(t) {ind_1, …, ind_10000})        // Calculate fitness
P_1(t) ← P_1(t) {ind_1, …, ind_10000}       // Sort individuals
P_1(t) ← P_1(t) {ind_1,… ind_1000}                   // Retain the 1000 best ones
                                   // Population P_2
P_2(t) ← {ind_1, …, ind_10000}              // Individuals of P_2 have neu+1 nodes in the hidden layer
f_2 (P_2(t) {ind_1, …, ind_10000}) ← fitness (P_2(t) {ind_1, …, ind_10000})        // Calculate fitness
P_2(t) ← P_2(t) {ind_1, …, ind_10000}       // Sort individuals
P_2(t) ← P_2(t) {ind_1,… ind_1000}                   // Retain the 1000 best ones
                                   // Evolution of populations P_1 and P_2 until 0.1*gen generations
for each P_i
  current_generation ← 0
    while current_generation < 0.1*gen not met do
      P_i(t) {ind_901,…ind_1000} ← P_i(t) {ind_1, …, ind_100}     // Best 10% replace the worst 10%
      P_i(t+1) ← P_i(t) {ind_1, …, ind_900}
      P_i(t+1) ← pm (P_i(t+1) {ind_1, …, ind_90})              // Parametric mutation (10% P_i (t+1))
      P_i(t+1) ← sm (P_i(t+1) {ind_91, …, ind_900})            // Structural mutation (90% P_i (t+1))
      f_i(P_i(t+1) {ind_1, .. ind_900}) ← fitness (P_i(t+1) {ind_1, …, ind_900})      // Evaluate
      P_i(t+1) ← P_i(t+1) (ind_1, …, ind_900) ∪ P_i(t){ind_901, …, ind_1000}
      P_i(t+1) ← P_i(t+1){ind_1, …, ind_1000}                  // Sort individuals
      current_generation ← current_generation + 1
    end while
end for
P(t) ← P_1{ind_1, …, ind_500} ∪ P_2{ind_1, …, ind_500}          // Individuals of P has [neu, neu+1]
                                                              // nodes in the hidden layer
P(t) ← P(t) {ind_1, …, ind_1000}                               // Sort individuals by fitness: ind_i > ind_i+1
                                   // Second Stage
result ← main_loop_Evolutionary Algorithm (Fig. 2, Steps 6-17) over P, Input: gen, neu+1
return result
```

**Fig. 4.** Pseudocode of the TSEA.

2007) and the last two concern complex real-world problems. The following fourteen have been used: Statlog (*Australian* credit approval), *Balance* scale, breast *Cancer* Wisconsin, *Heart* disease (Cleveland), *Hepatitis*, *Horse* colic, Thyroid disease (allhypo, *Hypothyroid*), *Ionos* (Ionosphere), *Liver* disorders, Thyroid disease (*Newthyroid*), *Pima* Indians diabetes and *Waveform* database generator (version 2) regarding the UCI data sets, and *BTX* and *Listeria monocytogenes* as real-world problems.

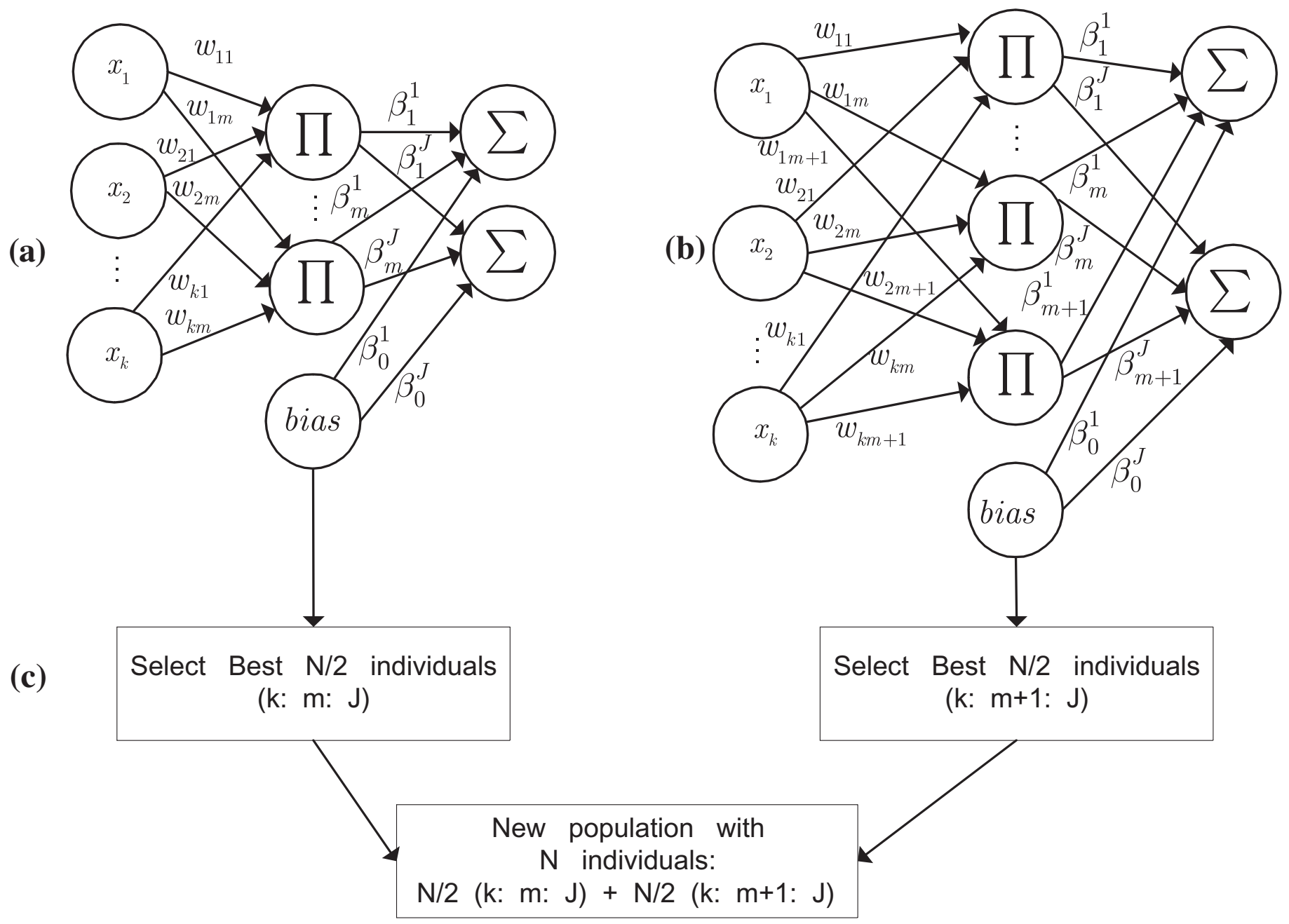

**Fig. 5.** Structure of a PUNN model with TSEA.

**Table 3**
Description of the TSEA configurations.

| Configuration | Num. of neurons in each population | Size of each population | Num. of generations in each population | $\alpha_2$ |
|---|---|---|---|---|
| 1* | neu and neu+1 | 1000 | 0.1**gen* | 1 |
| 2* | neu and neu+1 | 1000 | 0.1**gen* | 1.5 |

**Table 4**
Summary of the data sets used.

| Data set | Total patterns | Training patterns | Test patterns | Features | Inputs | Classes |
|---|---|---|---|---|---|---|
| Australian | 690 | 517 | 173 | 14 | 51 | 2 |
| Balance | 625 | 469 | 156 | 4 | 4 | 3 |
| Cancer | 699 | 525 | 174 | 10 | 9 | 2 |
| Heart | 303 | 227 | 76 | 13 | 26 | 2 |
| Hepatitis | 155 | 117 | 38 | 19 | 19 | 2 |
| Horse | 368 | 276 | 92 | 27 | 83 | 2 |
| Hypothyroid | 3772 | 2829 | 943 | 29 | 29 | 4 |
| Ionos | 351 | 263 | 88 | 34 | 34 | 2 |
| Liver | 345 | 259 | 86 | 6 | 6 | 2 |
| Newthyroid | 215 | 161 | 54 | 5 | 5 | 3 |
| Pima | 768 | 576 | 192 | 8 | 8 | 2 |
| Waveform | 5000 | 3750 | 1250 | 40 | 40 | 3 |
| BTX | 63 | 42 | 21 | 3 | 3 | 7 |
| Listeria | 539 | 305 | 234 | 4 | 4 | 2 |

BTX is a multi-class classification problem in the environment for different types of drinking waters (Hervás, Silva, Gutiérrez, & Serrano, 2008). The data set was obtained using a set of 63 drinking water samples spiked with individual standards of Benzene, Toluene or Xylene as well as with binary or ternary mixtures of them at concentrations between 5 and 30 μg/l, which constitutes an overall data set composed of seven different classes of contaminated drinking water samples with the same number of patterns.

*Listeria monocytogenes* is a bi-class problem in predictive microbiology. It has been a serious issue that has concerned food industries due to its ubiquity in the natural environment (Beuchat, 1996; Fenlon, Wilson, & Donachie, 1996) and the specific growth conditions of the pathogen that lead to its high prevalence in different kinds of food products. One impetus for this research was the problem of listeriosis (Tienungoon, Ratkowsky, McMeekin, & Ross, 2000), and different strategies were proposed to limit levels of contamination at the time of consumption to less than 100 CFU/g (European Commission, (Commission, 1999)). A fractional factorial design was followed in order to find out the growth limits of *Listeria monocytogenes*. Data were collected (Valero, Hervás, García-Gimeno, & Zurera, 2007) at citric and ascorbic acid concentrations between 0 and 0.4% (w/v) at intervals of 0.05%, at 4°, 7°, 10°, 15° and 30° C and pH levels of 4.5, 5, 5.5 and 6. This data set was divided so that 305 conditions covering the extreme domain of the model were chosen for training, and 234 conditions were selected within the range of the model to test its generalization capacity. Among the different conditions tested, there were 240 no-growth cases and 299 growth cases.

The size of the data sets ranges from almost one hundred to five thousand. The number of features depends on the problem and varies between three and forty, while the number of classes is between two and seven. The column labelled Inputs represents the number of input nodes in the PUNN model. Since we are using neural networks, all nominal variables have been converted to binary ones; due to this, sometimes the number of inputs is greater than the number of features. Also, the missing values have been replaced in the case of nominal variables by the mode or, when concerning continuous variables, by the mean, considering the full data set.

### 4.2. Validation technique and parameters employed

The experimental design uses the cross-validation technique called hold-out that consists of splitting the data into two sets: a training and a test set. The former is employed to train the neural network and the latter is used to test the training process and to measure neural network generalization capability. In our case, the size of the training set is $3n/4$ and that of the test set is approximately $n/4$, where $n$ is the number of patterns in the problem; these percentages are similar to those used in Prechelt (1994). We have employed a stratified holdout where the two sets are stratified (Kohavi, 1995) so that the class distribution of the samples in each set is approximately the same as in the original data set. The proportions do not match in Listeria because the data is prearranged in two sets due to their specific features.

The concrete values of the *neu* and *gen* parameters depend on the data set and are shown in Table 5. The decision about the number of neurons is a very difficult task in the scope of neural networks. The performance of the classifier might be better with other values, but determining the optimal values is a challenge. With respect to the number of generations, we have defined three kinds of values: small (100–150), medium (300) and large (500). Again, the optimal number is unknown; however the algorithm has a stop criterion to avoid evolving up to the maximum number of generations if there is no improvement. We have given values of our choice to the two parameters depending on the complexity of the data set (number of classes, inputs, instances, etc).

## 5. Results

First of all, this section presents the results obtained with respect to the CCR in the test set, $CCR_G$, with EDD and TSEA. As mentioned previously, the topologies are indicated as a sequence where the values will be separated by a colon. With TSEA, the number of nodes in the hidden layer for each data set belongs to one interval, given that the number of nodes will be different in each of the two populations that are combined. Thus, the second value will be an interval.

After that, a statistical analysis compares TSEA versus EDD. Next, the results of a second experiment are shown where other models of neural networks have been considered in order to determine the general performance of the PUNN.

**Table 5**
Values of the TSEA and/or EDD parameters depending on the data set.

| Data set | Num. of neurons (*neu*) | Num. of neurons in each population (*neu* and *neu*+1) | Max. Num. of generations (*gen*) | Num. of generations in each population |
|---|---|---|---|---|
| Australian | 4 | 4 and 5 | 100 | 10 |
| Balance | 5 | 5 and 6 | 150 | 15 |
| Cancer | 2 | 2 and 3 | 100 | 10 |
| Heart | 3 | 3 and 4 | 300 | 30 |
| Hepatitis | 3 | 3 and 4 | 100 | 10 |
| Horse | 4 | 4 and 5 | 300 | 30 |
| Hypothyroid | 3 | 3 and 4 | 500 | 50 |
| Ionos | 4 | 4 and 5 | 500 | 50 |
| Liver | 4 | 4 and 5 | 300 | 30 |
| Newthyroid | 3 | 3 and 4 | 300 | 30 |
| Pima | 3 | 3 and 4 | 120 | 12 |
| Waveform | 3 | 3 and 4 | 500 | 50 |
| BTX | 5 | 5 and 6 | 500 | 50 |
| Listeria | 4 | 4 and 5 | 300 | 30 |

Finally, we report a summary of the results obtained with a variety of classifiers, from the scope of neural networks or classical/modern machine learning. In most cases, it has been possible to conduct the experiments, so the partitions of the data sets are the same.

### 5.1. Results applying EDD and TSEA

The results obtained by applying the EDD (Tallón-Ballesteros et al., 2007) are presented, along with those obtained with TSEA. In the case of EDD, the parameters that are distributed throughout the processing nodes are the number of hidden-layer nodes, the maximum number of generations and the variance value ($\alpha_2$). There were 8 configurations, denoted in the following way: 1, 2 ... 8. As already mentioned, this paper only deals with the four configurations with the longest training process; the values of the parameters of each of them can be seen in Table 2, Section 3. In TSEA, the two existing configurations are 1* and 2*. The configurations 1 and 2 of EDD are equivalent to 1*, and analogously 3 and 4 are equivalent to 2*.

Table 6 shows the mean and standard deviation of the $\text{CCR}_G$ and the number of connections for each data set for a total of 30 runs or iterations. The best results regarding accuracy and complexity (measured in number of connections) appear in boldface for each data set. Table 6 depicts the generalization results obtained with EDD and TSEA methodologies. From the analysis of the data, it can be concluded, from a purely descriptive point of view, that the TSEA method obtains the best result in mean of the $\text{CCR}_G$ for all data sets. The differences have to be considered to be between the mean value of $\text{CCR}_G$ obtained with TSEA and the mean values of the two equivalent configurations in EDD. For instance, in Balance the results obtained with configuration 1* with 5–6 neurons surpass configurations 1 and 2 with populations of only 5 or 6 neurons. The mean number of connections often increases in TSEA configurations, corresponding to those where improvements in $\text{CCR}_G$ results are produced with respect to equivalent configurations.

Furthermore, another advantage is that approximately 40% fewer evaluations per iteration are needed with TSEA to get results comparable to those previously obtained with EDD. Now in this joint experiment, there are two types of individuals in the same population, with *neu* and *neu+1* nodes without duplicating the total size of the population; in the past, these two experiments were

**Table 6**
Results obtained in fourteen data sets applying EDD and TSEA.

| Data set | Configuration (Topology) | $\text{CCR}_G$ Mean ± SD | Mean Num. of Conn. ± SD | Data set | Configuration (Topology) | $\text{CCR}_G$ Mean ± SD | Mean Num. of Conn. ± SD |
|---|---|---|---|---|---|---|---|
| Australian | 1 (51:4:1) | 87.63 ± 1.49 | 49.93 ± 13.97 | Ionos | 1 (34:4:1) | 92.31 ± 2.23 | 39.63 ± 7.53 |
| | 2 (51:5:1) | 87.32 ± 1.66 | 52.47 ± 13.36 | | 2 (34:5:1) | 92.50 ± 2.28 | 49.37 ± 8.97 |
| | 1* (51:[4,5]:1) | 88.11 ± 1.56 | 43.50 ± 15.34 | | 1* (34:[4,5]:1) | 91.51 ± 1.88 | **33.67** ± **6.45** |
| | 3 (51:4:1) | 87.70 ± 1.45 | 49.83 ± 13.27 | | 3 (34:4:1) | 91.85 ± 2.48 | 37.53 ± 7.21 |
| | 4 (51:5:1) | 87.63 ± 1.54 | 53.90 ± 16.59 | | 4 (34:5:1) | 93.06 ± 1.87 | 48.90 ± 9.74 |
| | 2* (51:[4,5]:1) | **88.68** ± **1.10** | **38.00** ± **11.52** | | 2* (34:[4,5]:1) | **93.22** ± **1.92** | 38.80 ± 7.23 |
| Balance | 1 (4:5:2) | 95.30 ± 1.47 | **22.87** ± **2.52** | Liver | 1 (6:4:1) | 73.87 ± 2.21 | 18.43 ± 1.94 |
| | 2 (4:6:2) | 95.15 ± 0.98 | 25.23 ± 2.22 | | 2 (6:5:1) | 73.56 ± 1.97 | 21.47 ± 2.85 |
| | 1* (4:[5,6]:2) | **96.20** ± **1.06** | 24.83 ± 2.36 | | 1* (6:[4,5]:1) | **74.61** ± **2.00** | 19.83 ± 2.15 |
| | 3 (4:5:2) | 95.04 ± 1.41 | 23.37 ± 3.06 | | 3 (6:4:1) | 72.20 ± 3.22 | **17.37** ± **2.47** |
| | 4 (4:6:2) | 95.62 ± 1.16 | 26.67 ± 3.29 | | 4 (6:5:1) | 73.64 ± 3.23 | 22.17 ± 2.59 |
| | 2* (4:[5,6]:2) | 95.62 ± 1.12 | 25.87 ± 3.04 | | 2* (6:[4,5]:1) | 72.36 ± 1.51 | 20.23 ± 1.38 |
| Cancer | 1 (9:2:1) | 98.49 ± 0.61 | 12.23 ± 1.50 | Newthyroid | 1 (5:3:2) | 94.44 ± 0.68 | 16.97 ± 2,58 |
| | 2 (9:3:1) | 98.97 ± 0.38 | 15.80 ± 1.73 | | 2 (5:4:2) | 94.75 ± 0.98 | 21.80 ± 2.92 |
| | 1* (9:[2,3]:1) | 98.74 ± 0.61 | 16.40 ± 2.44 | | 1* (5:[3,4]:2) | **94.88** ± **0.93** | 20.10 ± 2.28 |
| | 3 (9:2:1) | 98.51 ± 0.49 | **11.97** ± **1.79** | | 3 (5:3:2) | 94.81 ± 0.89 | 17.30 ± 2.56 |
| | 4 (9:3:1) | 98.72 ± 0.60 | 16.23 ± 2.49 | | 4 (5:4:2) | 94.75 ± 1.38 | 22.77 ± 3.00 |
| | 2* (9:[2,3]:1) | **98.98** ± **0.54** | 15.90 ± 2.01 | | 2* (5:[3,4]:2) | 94.19 ± 1.92 | **15.37** ± **1.97** |
| Heart | 1 (26:3:1) | 82.18 ± 2.36 | 33.47 ± 6.97 | Pima | 1 (8:3:1) | 77.33 ± 2.36 | **13.53** ± **2.39** |
| | 2 (26:4:1) | 82.36 ± 2.98 | 40.13 ± 7.99 | | 2 (8:4:1) | 78.61 ± 1.88 | 18.60 ± 3.04 |
| | 1* (26:[3,4]:1) | **83.68** ± **2.57** | 41.77 ± 8.73 | | 1* (8:[3,4]:1) | **78.63** ± **1.33** | 13.73 ± 2.77 |
| | 3 (26:3:1) | 82.32 ± 3.08 | **30.67** ± **8.73** | | 3 (8:3:1) | 76.96 ± 1.67 | 13.73 ± 1.91 |
| | 4 (26:4:1) | 81.97 ± 3.05 | 44.33 ± 10.05 | | 4 (8:4:1) | 77.69 ± 1.79 | 17.37 ± 2.20 |
| | 2* (26:[3,4]:1) | 83.64 ± 2.33 | 45.67 ± 6.15 | | 2* (8:[3,4]:1) | 77.69 ± 1.33 | 15.83 ± 2.59 |
| Hepatitis | 1 (19:3:1) | 84.47 ± 4.49 | 27.23 ± 4.53 | Waveform | 1 (40:3:2) | 81.43 ± 2.10 | **30.75** ± **6.73** |
| | 2 (19:4:1) | 85.52 ± 4.67 | 35.90 ± 5.21 | | 2 (40:4:2) | 82.78 ± 0.64 | 48.63 ± 9.36 |
| | 1* (19:[3,4]:1) | **85.79** ± **4.51** | 31.67 ± 5.63 | | 1* (40:[3,4]:2) | **84.46** ± **0.92** | 43.13 ± 5.72 |
| | 3 (19:3:1) | 84.47 ± 4.55 | **26.80** ± **4.01** | | 3 (40:3:2) | 82.05 ± 1.64 | 36.25 ± 7.78 |
| | 4 (19:4:1) | 84.29 ± 5.33 | 36.53 ± 6.65 | | 4 (40:4:2) | 84.32 ± 1.73 | 43.25 ± 5.73 |
| | 2* (19:[3,4]:1) | 83.68 ± 3.87 | 33.63 ± 3.93 | | 2* (40:[3,4]:2) | 82.01 ± 1.48 | 40.88 ± 12.86 |
| Horse | 1 (83:4:1) | 86.41 ± 2.38 | **82.60** ± **15.38** | BTX | 1 (3:5:6) | 79.04 ± 6.32 | 39.53 ± 3.41 |
| | 2 (83:5:1) | 85.18 ± 2.57 | 108.10 ± 24.13 | | 2 (3:6:6) | 77.61 ± 7.09 | 43.57 ± 3.77 |
| | 1* (83:[4,5]:1) | 85.50 ± 2.97 | 89.70 ± 27.35 | | 1* (3:[5,6]:6) | 79.68 ± 7.39 | **38.27** ± **5.17** |
| | 3 (83:4:1) | 85.72 ± 2.43 | 84.67 ± 22.05 | | 3 (3:5:6) | 78.73 ± 5.26 | 39.47 ± 3.57 |
| | 4 (83:5:1) | 85.57 ± 3.60 | 107.97 ± 20.32 | | 4 (3:6:6) | 77.14 ± 4.90 | 44.90 ± 3.66 |
| | 2* (83:[4,5]:1) | **86.59** ± **2.38** | 99.17 ± 25.15 | | 2* (3:[5,6]:6) | **81.11** ± **6.55** | 38.93 ± 3.79 |
| Hypothyroid | 1 (29:3:3) | 95.27 ± 0.77 | **31.13** ± **6.38** | Listeria | 1 (4:4:1) | 87.20 ± 1.71 | 14.93 ± 1.48 |
| | 2 (29:4:3) | 95.32 ± 0.58 | 39.63 ± 8.78 | | 2 (4:5:1) | 87.45 ± 1.14 | 18.63 ± 1.73 |
| | 1* (29:[3,4]:3) | **95.37** ± **0.40** | 42.00 ± 7.41 | | 1* (4:[4,5]:1) | 86.54 ± 1.67 | **14.63** ± **1.73** |
| | 3 (29:3:3) | 94.96 ± 0.62 | 34.38 ± 2.92 | | 3 (4:4:1) | 86.66 ± 1.82 | 15.03 ± 1.59 |
| | 4 (29:4:3) | 95.16 ± 0.32 | 33.50 ± 7.65 | | 4 (4:5:1) | 86.98 ± 1.72 | 18.33 ± 1.92 |
| | 2* (29:[3,4]:3) | 94.94 ± 0.25 | 43.00 ± 8.45 | | 2* (4:[4,5]:1) | **87.68** ± **1.06** | 17.43 ± 1.52 |

carried out separately. The mean number of evaluations per iteration with EDD is given by:

$$\text{evalutions(EDD)} = pop_size^{*}10 + 0.9^{*}pop_size^{*}gen, \tag{7}$$

where *pop_size* is the population size and *gen* the maximum number of generations.

On the other hand, with TSEA the expression is the following:

$$\text{evalutions(TSEA)} = (pop_size^{*}10 + 0.9^{*}pop_size^{*}0.1^{*}gen)^{*}2 + 0.9^{*}pop_size^{*}gen. \tag{8}$$

As stated before, only one experiment with TSEA is needed, while two are necessary with EDD. Table 7 shows the mean numerical values of the number of evaluations per iteration for all the data sets concerned in the single TSEA experiment and the two EDD ones, which are equivalent, along with the reduction percentage in the number of evaluations. This percentage ranges from 36% to 39%, depending on the number of generations for each data set.

### 5.2. Statistical analysis

In this section we will perform an ANalysis Of VAriance (ANOVA) for the $CCR_G$, for each data set and determine if there are significant differences between the numbers of nodes considered or the values of $\alpha_2$ for the underlying models. These two factors are not independent so an analysis will be carried out using all possible pairs $(N,A)$ of the Cartesian product of the two sets. For each pair, or cell, 30 runs of the EA have been performed with different random seeds. Table 6 shows the mean and standard deviation values obtained in these runs.

#### 5.2.1. Accuracy and complexity analysis

First of all, hypothesis tests are performed to try to determine the mean effect of each term on the $CCR_G$ of the best individuals in the last generation for each run. Tests have been carried out for every factor and for the interaction among factors. A normal distribution can be assumed for all the variables contrasted, because the *p*-values of the Kolmogorov–Smirnov (K–S) test are over a significance coefficient of 0.05. Thus, the results have been studied by means of an analysis of variance ANOVA II (Dunn & Clark, 1974; Miller, 1981; Snedecor & Cochran, 1980) with the $CCR_G$ of the best individuals for each run, and CCR as the test variable. This CCR is obtained independently in 30 runs and depends on two fixed factors and their interaction. The linear model has the form:

$$CCR_{ijk} = \mu + N_i + A_j + NA_{ij} + \varepsilon_{ijk} \quad \text{for } i = 1,2,3;\ j = 1,2 \quad \text{and} \quad k = 1,\ldots,30, \tag{9}$$

where the parameter $\mu$ is the global mean of the model. $N_i$ is the effect on CCR of the *i*-th level of factor *N*, the number of nodes in the hidden layer, where $N_1$ = neu, $N_2$ = neu+1 and $N_3$ = (neu, neu+1); $A_j$ is the effect on CCR of the *j*-th level of factor $\alpha_2$, where $A_1 = 1$ and $A_2 = 1.5$; $NA_{ij}$ represents the effect of the interaction between different values of the number of nodes and the $\alpha_2$. Finally, $\varepsilon_{ijk}$ are error variables associated with effects on the CCR, other factors not observed in the experiment and, for those not present in the linear model and/or observation error, measure error, etc. The variation in experimental results from CCR is explained by the effects of different levels in the factors of the model and their interaction.

In a second step, if there are significant differences in mean $CCR_G$, a comparison test of $CCR_G$ means will be performed for each data set in order to see if the TSEA methodology supplies significantly better values in accuracy (measured in CCR) than EDD.

For the A factor, associated with the $\alpha_2$ value, a student's *t*-test was performed in order to ascertain whether the differences in CCR mean between the two different values of $\alpha_2$ considered were significant. For factor *N*, there is a multiple comparison test of the average CCR obtained with the three different levels to determine whether there are significant differences. Thus, 90 simulations were done, corresponding to the 30 runs of each level.

The *p*-values, $p^*$, in Table 8 of each term of the linear model show that the number of nodes in the hidden layer significantly affects the $CCR_G$ mean in seven data sets at a significance level of 5%, for instance, in the Heart data set *p*-value = 0.008 (see the second column in Table 8). If now the effect of the *A* factor is taken into consideration, it can be inferred that for the Liver and Pima data sets, there exists a significant effect in the mean of CCR based on $\alpha_2$ values (*p*-values 0.035 and 0.001 are lower than 0.050). With respect to interactions among the factors, the fourth column in Table 8, NA, shows that only for the Liver, Waveform and Listeria data sets does a significant effect exist in the mean of the CCR based on that interaction (*p*-values 0.028, 0.001 and 0.016 are lower than 0.050). In the remaining data sets, the effect of this interaction has been added to the error term and the Snedecor's F tests have been redone (this situation is indicated in Table 8 with a – sign).

Regarding the complexity, a similar analysis of variance ANOVA II with the number of connections has been done. The *p*-values, $p^* < 0.050$, of factor *N* in the linear model (see Table 8) indicate that there are significant differences between the mean values of the number of connections for each architecture used for each data set, whereas there are not significant differences in the average of connections depending on the parameter $\alpha_2$ except for the Balance, Newthyroid and Listeria data sets (*p*-values are 0.018, 0.003 and 0.001). As for the interaction between the two factors, it is significant for the Heart, Horse, Newthyroid, Pima and Listeria data sets, which indicate that $N$ and $\alpha_2$ values are affected when these act jointly on the mean of the number of connections of network models.

**Table 7**
Mean number of evaluations per iteration with TSEA and EDD and percentage of reduction.

| Data set | TSEA | EDD | Reduction (%) |
|---|---|---|---|
| Australian | 128000 | 200000 | 36 |
| Balance | 182000 | 290000 | 37 |
| Cancer | 128000 | 200000 | 36 |
| Heart | 344000 | 560000 | 39 |
| Hepatitis | 128000 | 200000 | 36 |
| Horse | 344000 | 560000 | 39 |
| Hypothyroid | 560000 | 920000 | 39 |
| Ionos | 560000 | 920000 | 39 |
| Liver | 344000 | 560000 | 39 |
| Newthyroid | 344000 | 560000 | 39 |
| Pima | 149600 | 236000 | 37 |
| Waveform | 560000 | 920000 | 39 |
| BTX | 560000 | 920000 | 39 |
| Listeria | 344000 | 560000 | 39 |

**Table 8**
*p*-Values of the F test of the ANOVA II methodology for the means of the $CCR_G$ and the number of connections.

| Data set | $CCR_G$ F test | | | Num. of connections F test | | |
|---|---|---|---|---|---|---|
| | *N* | *A* | *NA* | *N* | *A* | *NA* |
| Australian | 0.002 | 0.147 | – | 0.000 | 0.510 | – |
| Balance | 0.004 | 0.508 | – | 0.000 | 0.018 | – |
| Cancer | 0.001 | 0.694 | – | 0.000 | 0.713 | – |
| Heart | 0.008 | 0.569 | – | 0.000 | 0.150 | 0.033 |
| Hepatitis | 0.831 | 0.171 | – | 0.000 | 0.342 | – |
| Horse | 0.394 | 0.954 | – | 0.000 | 0.264 | 0.033 |
| Hypothyroid | 0.662 | 0.085 | – | 0.002 | 0.769 | |
| Ionos | 0.098 | 0.257 | – | 0.000 | 0.468 | – |
| Liver | 0.417 | 0.001 | 0.028 | 0.000 | 0.974 | – |
| Newthyroid | 0.536 | 0.648 | – | 0.000 | 0.003 | 0.000 |
| Pima | 0.006 | 0.035 | – | 0.000 | 0.343 | 0.002 |
| Waveform | 0.003 | 0.825 | 0.001 | 0.001 | 0.776 | – |
| BTX | 0.035 | 0.823 | – | 0.000 | 0.182 | – |
| Listeria | 0.591 | 0.916 | 0.016 | 0.000 | 0.001 | 0.000 |

Generally speaking, we can conclude that the TSEA methodology improves the CCR values throughout the test set in the majority of the data sets and, for a great deal of them, quite significantly. The number of connections increases in a significant way.

### *5.3. Product versus sigmoidal units and radial basis function neural networks*

This section involves a comparison in performance (measured in $CCR_G$) between neural networks based on product units, sigmoidal units and radial basis functions. For this comparison four methodologies have been used: the traditional MLP model (Bishop, 1995) with a learning Back-Propagation method (BP); the RBF model (Howlett & Jain, 2001); the PUs with EDD methodology (EDD); and the PUs with TSEA methodology (TSEA). These methods were run with all the data sets in question. We have used the Weka tool (Witten & Frank, 2005) to run algorithms BP and RBF.

The parameters for BP were the following: learning rate $\eta = 0.3$ and momentum $\alpha = 0.2$. The number of epochs was adjusted in each data set. Regarding the topology of the models (number of hidden nodes), in the case of MLP and RBF we have considered default topology. In EDD and TSEA we deal with the best topology that we have reported in this paper. The number of runs for MLP and RBF was 30.

Table 9 includes the results of TSEA, EDD, MLP and RBF methodologies for all data sets. The best results in $CCR_G$ for each data set are shown in boldface.

From a purely descriptive analysis of the results, it can be concluded that the TSEA method obtains the best result for nine data sets, the RBF methodology yields the highest performance for four data sets, whereas MLP only does so for one data set. Furthermore, the TSEA method obtains the best mean ranking ($\overline{R} = 1.57$), followed by the EDD method ($\overline{R} = 2.64$), and reports the highest mean accuracy ($\overline{CCR_G} = 87.85\%$) followed by the EDD method ($\overline{CCR_G} = 87.38\%$). The mean accuracy and ranking obtained by the TSEA and EDD are higher than those obtained by MLP and RBF, which confirms the best accuracy of the product units.

To determine the statistical significance of the differences in rank observed for each method in the different data sets, a non-parametric Friedman test (Friedman, 1937) has been carried out with the $CCR_G$ ranking of the best models as the test variable (since a previous evaluation of the $CCR_G$ values results in rejecting the normality and the equality of the variances hypothesis). The null-hypothesis states that all algorithms are equivalent and so their ranks ($\overline{R}i$) should be equal. The test shows that the effect of the method used for classification is statistically significant at a significance level of 5%, as the confidence interval is $C_0 = (0, F_{0.05} = 2.84)$ with a F-distribution with 3 and 39 degrees of freedom, and the F-distribution statistical value (Iman & Davenport, 1980) is $F^* = 4.08 \notin C_0$. Consequently, we reject the null-hypothesis; thus it implies the existence of significant differences in the performances of all the methods considered. On the basis of this rejection, a post-hoc non-parametric Bonferroni–Dunn test (Hochberg & Tamhane, 1987) was applied with the best performing algorithm TSEA as the control method. The results of the Bonferroni–Dunn test for $\alpha = 0.1$ and $\alpha = 0.05$ can be seen in Table 10 using the corresponding critical values for the two-tailed Bonferroni–Dunn test.

The TSEA method obtains a significantly higher ranking of $CCR_G$ when compared to all the methods and none of the remaining methods obtain significantly higher rankings when compared to each other. Consequently, the TSEA method obtains a significantly higher performance than the three methods, EDD, MLP and RBF, which justifies the proposal.

Therefore, we can conclude that the results obtained by TSEA make it a very competitive method when compared to the other neural network learning schemes previously mentioned.

### *5.4. Results obtained with a variety of classifiers*

Now, a general review can be made of the results obtained with another kind of neural networks and other machine learning algorithms. In the literature a huge amount of tests have been carried out with some of the data sets here considered. However, in some cases the classifier attribute is different, the number of instances is not the same, since some have been removed, or even the number of features does not match; the result is not included for those situations. The method of cross-validation is different in many papers, so the comparison is not as fair as we would like. Our purpose is to view some of the methods that have been tested with

**Table 10**
Critical difference values, mean ranking and differences of rankings of the Bonferroni–Dunn test, using TSEA as the control method.

| Method | $\overline{R}$ | Difference with TSEA |
|---|---|---|
| TSEA | $\overline{R}_{(1)} = 1.57$ | – |
| EDD | $\overline{R}_{(2)} = 2.64$ | $\lvert\overline{R}_{(2)} - \overline{R}_{(1)}\rvert = 1.07^{a}$ |
| MLP | $\overline{R}_{(3)} = 2.86$ | $\lvert\overline{R}_{(3)} - \overline{R}_{(1)}\rvert = 1.29^{b}$ |
| RBF | $\overline{R}_{(4)} = 2.93$ | $\lvert\overline{R}_{(4)} - \overline{R}_{(1)}\rvert = 1.36^{b}$ |

CD($\alpha = 0.1$) = 1.04; CD($\alpha = 0.05$) = 1.17
(1): TSEA; (2): EDD; (3): RBF; (4): MLP.
CD: Critical Difference.
[a] Statistically significant difference with $\alpha = 0.1$.
[b] Statistically significant difference with $\alpha = 0.05$.

**Table 9**
Results of the comparison with other techniques.

| Method | Data set | | | | | | |
|---|---|---|---|---|---|---|---|
| | Australian | Balance | Cancer | Heart | Hepatitis | Horse | Hypothyroid |
| TSEA | **88.68 ± 1.10** | **96.20 ± 1.06** | **98.98 ± 0.54** | 83.68 ± 2.57 | 85.79 ± 4.51 | 86.59 ± 2.22 | **95.37 ± 0.40** |
| EDD | 87.70 ± 1.45 | 95.62 ± 1.16 | 98.97 ± 0.38 | 82.36 ± 2.98 | 85.52 ± 4.67 | 86.41 ± 2.38 | 95.32 ± 0.58 |
| MLP | 84.10 ± 1.48 | 93.78 ± 1.81 | 97.81 ± 0.47 | 84.82 ± 2.55 | 84.73 ± 2.08 | **88.51 ± 1.57** | 94.39 ± 0.32 |
| RBF | 75.84 ± 2.04 | 88.27 ± 1.83 | 97.20 ± 0.25 | **86.75 ± 2.39** | **89.30 ± 2.29** | 80.47 ± 1.38 | 92.83 ± 0.56 |
| | Ionos | Liver | Newthyroid | Pima | Waveform | BTX | Listeria |
| TSEA | **93.22 ± 1.92** | **74.61 ± 2.00** | 94.88 ± 0.93 | **78.63 ± 1.33** | 84.46 ± 0.92 | **81.11 ± 6.55** | **87.68 ± 1.06** |
| EDD | 93.06 ± 1.87 | 73.87 ± 2.21 | 94.81 ± 0.90 | 78.61 ± 1.88 | 84.32 ± 1.73 | 79.04 ± 6.32 | 87.45 ± 1.14 |
| MLP | 89.12 ± 1.54 | 65.65 ± 3.35 | 97.08 ± 1.40 | 75.94 ± 2.13 | 84.85 ± 0.96 | 54.12 ± 8.10 | 84.49 ± 1.70 |
| RBF | 92.46 ± 0.70 | 57.17 ± 3.43 | **98.27 ± 0.47** | 77.34 ± 2.17 | **87.29 ± 0.07** | 80.95 ± 0.00 | 83.70 ± 0.00 |

$\overline{CCR_G}(\text{TSEA}) = 87.85$; $\overline{CCR_G}(\text{EDD}) = 87.38$; $\overline{CCR_G}(\text{MLP}) = 84.24$; $\overline{CCR_G}(\text{RBF}) = 84.85$

**Table 11**
Summary of the results in fourteen data sets comparing TSEA to existing works related to neural networks or classical/modern machine learning approaches.

| Method | Australian | Balance | Cancer | Heart | Hepatitis | Horse | Hypothyroid |
|---|---|---|---|---|---|---|---|
| TSEA | **88.68** | **96.20** | **98.98** | 83.68 | 85.79 | 86.59 | 95.37 |
| EDD | 87.70 | *95.62* | *98.97* | 82.36 | 85.52 | 86.41 | 95.32 |
| HMOEN_HN | – | – | 96.82 | – | 75.51 | – | – |
| HMOEN_L2 | – | – | 96.30 | – | 80.30 | – | – |
| MLP | 84.10 | 93.78 | 97.81 | *84.82* | 84.73 | **88.51** | 94.39 |
| RBF | 75.84 | 88.27 | 97.20 | **86.75** | *89.30* | 80.47 | 92.83 |
| SONG | – | 87.80 | 97.40 | – | – | – | – |
| C4.5 | 86.71 | 83.33 | 97.13 | 75.00 | 84.21 | *88.04* | **99.15** |
| k-NN | 85.55 | 91.67 | 98.85 | 82.89 | 86.84 | *88.04* | 94.06 |
| PART | 84.97 | 85.26 | 97.13 | 80.26 | 81.58 | 85.87 | *98.83* |
| SVM | *88.44* | 88.46 | 98.28 | 82.89 | **89.47** | *88.04* | 93.85 |
| | Ionos | Liver | Newthyroid | Pima | Waveform | BTX | Listeria |
| TSEA | 93.22 | **74.61** | 94.88 | **78.63** | 84.46 | **81.11** | **87.68** |
| EDD | *93.06* | *73.87* | 94.81 | *78.61* | 84.32 | 79.04 | *87.45* |
| HMOEN_HN | – | 68.94 | – | 75.36 | – | – | – |
| HMOEN_L2 | – | 68.00 | – | 78.50 | – | – | – |
| MLP | 89.12 | 65.65 | 97.08 | 75.94 | 84.85 | 54.12 | 84.49 |
| RBF | 92.46 | 57.17 | **98.27** | 77.34 | *87.29* | *80.95* | 83.70 |
| SONG | 91.20 | 68.50 | 97.20 | 76.40 | – | – | – |
| C4.5 | 92.05 | 68.60 | 96.30 | 74.48 | 76.40 | *80.95* | 85.93 |
| k-NN | 90.91 | 63.95 | 94.44 | 75.00 | 81.12 | 76.19 | 85.93 |
| PART | **95.45** | 61.63 | 92.59 | 74.48 | 78.16 | *80.95* | 86.67 |
| SVM | 88.64 | 58.14 | 88.89 | 78.13 | **88.80** | 61.90 | 80.74 |

$\overline{CCR_G}(\mathrm{TSEA}) = 87.85$; $\overline{CCR_G}(\mathrm{EDD}) = 87.38$; $\overline{CCR_G}(\mathrm{MLP}) = 84.24$; $\overline{CCR_G}(\mathrm{RBF}) = 84.85$
$\overline{CCR_G}(\mathrm{C4.5}) = 84.88$; $\overline{CCR_G}(\mathrm{k-NN}) = 85.39$; $\overline{CCR_G}(\mathrm{PART}) = 84.56$; $\overline{CCR_G}(\mathrm{SVM}) = 83.91$

some of the data sets dealt with in the current paper. Regarding neural networks, only those methods using the same number of inputs as ours are considered. These methods are TSEA, EDD, HMOEN_HN (Goh, Teoh, & Tan, 2008), HMOEN_L2 (Goh et al., 2008), MLP, RBF and SONG (Inoue & Narihisa, 2005). Other classical or modern machine learning algorithms have been included: C4.5 (Quinlan, 1993), k-nearest neighbours (k-NN) (Aha, Kibler, & Albert, 1991; Cover & Hart, 1967), PART (Frank & Witten, 1998) and SVM (Vapnik, 1995). Since, MLP, RBF, C4.5, k-NN, PART and SVM are implemented in Weka tool (Witten & Frank, 2005), the same cross-validation is used as in the current proposal and in our previous work, thus the same instances in each of the partitions; regarding the parameters, the algorithms have been run with the default values. We have reported the best accuracy in k-NN where $k$ is 1, 3, 5, 7 or 9. For the remaining methods cited above, it is not possible to conduct the experiments, because the implementations are not publicly available. There does not exist one method that performs really well with all data sets; depending on the data set, the best classifier belongs to either the neural networks approach or to the classical/modern machine learning. In Table 11 the results are summarized; the best ones in boldface and in italics, the second best ones, as well as the averages of the methods run by us with all data sets.

## 6. Conclusions

The aim of this paper is to tackle multi-classification problems using neural networks based on product units, but at a lower computational cost than that used in algorithms proposed in previous works. Our basic assumption is that it is good to employ a methodology based on a population with more diverse models regarding network architectures and this produces an improvement in efficiency, maintaining good results.

The TSEA is applied to solve fourteen classification problems, twelve from the UCI repository and two real-world problems, with a great deal of variety in the number of instances, features and classes. The results confirm that our approach obtains promising results, achieving a high classification rate level in the data sets at a lower computational cost.

Experimental studies have been performed to determine if different values in the number of nodes in the hidden layer and $\alpha_2$ significantly affect performance.

The EA has been run with PUs using, on one hand, the standard EA applying EDD, and on the other hand, TSEA, a method that significantly reduces the computational cost measured by means of the number of EA evaluations. Once we have performed the statistical analysis (see Table 8), we conclude that TSEA maintains enough robust $CCR_G$ values and, in a considerable number of data sets, significantly higher ones.

In order to analyze performance with respect to efficacy, these two methodologies for the generation of PUNN models have been compared to other network models. The former have sigmoidal basis units and it is a model trained with a BP algorithm (MLP) whereas the latter is model trained with RBF networks.

According to the above results, our learning methodology of neural networks, TSEA, is seen to considerably improve efficiency and significantly efficacy in most of the data sets with respect to the EDD methodology. TSEA obtains a significantly better ranking compared to EDD, MLP and RBF. We have also summarized the results obtained with other kinds of neural networks and classical/modern machine learning algorithms.

## Acknowledgments

This work has been partially subsidized by TIN2007-68084-C02-02 and TIN2008-06681-C06-03 projects of the Spanish Inter-Ministerial Commission of Science and Technology (MICYT), FEDER funds and the P08-TIC-3745 project of the "Junta de Andalucía" (Spain).